# A Letter on Convergence of In-Parameter-Linear Nonlinear Neural Architectures with Gradient Learnings

Ivo Bukovsky, *Senior, Member, IEEE,* Gejza Dohnal, Peter M. Benes, Kei Ichiji, & Noriyasu Homma*, Member, IEEE*

*Abstract* — This letter summarizes and proves the concept of bounded-input bounded-state (BIBS) stability for weight convergence of a broad family of in-parameter-linear nonlinear neural architectures as it generally applies to a broad family of incremental gradient learning algorithms. A practical BIBS convergence condition results from the derived proofs for every individual learning point or batches for real-time applications.

## I. Introduction

THE bounded-input bounded-state (BIBS) stability concept is recently popular in neural networks. Also, the weight convergence of gradient learning is still an investigated issue. However, no paper recalls nor thoroughly presents and prooves this concept for the incremental gradient-learning weight convergence of in-parameter-linear nonlinear (neural) architectures (IPLNAs) in general.

By IPLNAs in this paper, we consider a wide family of shallow neural networks including the extreme learning machine (ELM) or random vector functional link (RVFL) networks [1]–[3], other functional links and kernel neural architectures and filters, e.g., [4]–[6], and polynomial neural networks [7], [8] including basic standalone ΣΠ structures called polynomial neural units in [9], [10], and references therein.

This letter shows that the input-to-state stability (ISS) concept [11] and BIBS stability [12] generally apply to the gradient learning algorithms and their many modifications for IPLNA's. The most prevalent ones to mention are the (stochastic) gradient descent with normalized learning rate (also as normalized least means squares (NLMS) [13]–[15]), recursive least squares (RLS) [16], adaptive moment estimation (ADAM) [17].

In general, there are many modifications of gradient learnings that are based on:
- the normalization of learning rate and the adaptation of the regularization term (such as NLMS, generalized gradient descent (GNGD) [18], robust regularized NLMS (RR-NLMS) [19])
- the adaptation of the learning rate (such as Benveniste's [13], Farhang's & Ang's [20], Mathew's [21]), and
- other variations, e.g., RLS that uses covariance matrix and momentum enhanced methods such as ADAM and the related predecessors and followers as overviewed in [22].

The purpose of this paper is to summarize and prove the general applicability of ISS and BIBS to the weight convergence of IPLNAs and to highlight its practical aspect for the broad family of gradient learning rules of IPLNAs.

The bold letters and symbols stand for vectors and matrices, and sample index $k$ indicates time variability.

## II. Background on IPLNAs and Gradient Learnings

This subsection generally defines IPLNAs and recalls BIBS stability concept applied to their gradient-based weight-update system.

[1]vo Bukovsky acknowledges support by the FSP JU (Fund of Strategic Priorities of the University of South Bohemia in Ceske Budejovice).
Gejza Dohnal and Ivo Bukovsky acknowledges support from the ESIF, EU Operational Programme Research, Development and Education, and from the Center of Advanced Aerospace Technology (CZ.02.1.01/0.0/0.0/16\_019/0000826), Faculty of Mechanical Engineering, Czech Technical University in Prague.
Noriyasu Homma acknowledges support by the JSPS KAKENHI 18K19892 and 19H04479, and the Smart Aging Research Center, Tohoku University.

Ivo Bukovsky is with the Dpt. of Computer Science, Faculty of Science, University of South Bohemia in Ceske Budejovice, and in part with the Dpt. of Mechaniccs, Biomechanics, and Mechatronics, Faculty of Mechanical Engineering, Czech Technical University in Prague, Czech Republic (corresponding author, email: ibukovsky@ieee.org).
Gejza Dohnal is with the Dpt. of Technical Mathematics, Faculty of Mechanical Engineering, Czech Technical University in Prague.
Noriyasu Homma is with the Tohoku University Graduate School of Medicine, 2-1 Seriyo-machi, Aoba-ku, Sendai, 980-8575, Japan.

***Definition 1:*** The in-parameter-linear nonlinear neural architecture (IPLNA) is defined by (1) where $\mathbf{g}(\mathbf{x})$ is not a function of neural weights $\mathbf{w}$, i.e., $\mathbf{g}(\mathbf{x}(\mathbf{v},k)) \neq \mathbf{g}(\mathbf{w})$.

The IPLNA that conforms to Definition 1 is as follows

$$\tilde{y}(k) = \mathbf{w}^T(k) \cdot \mathbf{g}(\mathbf{x}(\mathbf{v},k)), \quad (1)$$

where $\tilde{y}(k)$ is neural output; $\mathbf{w}$ is the column vector of neural weights; upper $^T$ stand for transposition; $\mathbf{g}()$ can be some

- vector of basis functions or kernel vector function, or
- random vector functional link expansion as with ELMs and RVFLs;

i.e., $\mathbf{g}(\mathbf{x})$ transforms the basic feature vector $\mathbf{x}$ into a new feature vector $\mathbf{g}(\mathbf{x})$ independently of $\mathbf{w}$ (Definition 1). Further in (1), $\mathbf{v}=\mathbf{v}(k)$ are additional parameters that can vary in time. The gradient learning rule details for IPLNAs (1) and some of their distinctions for variations of gradient learning algorithms are sketched in Tab. 1, where most distinctions lie in the time-varying learning rate $\eta=\eta(k)$. For notational simplicity, $\mathbf{g}(\mathbf{x}(\mathbf{v},k))$ can be shortened as $\mathbf{g}(\mathbf{x})$ or $\mathbf{g}(\mathbf{x},k)$ and so on.

**Tab. 1:** General incremental gradient-learning scheme for in-parameter-linear nonlinear neural architectures (IPLNAs) and its three most popular incremental gradient learning variations (briefly sketched to indicate the principal differences while the details can be found in cited literature and references therein)

| general gradient-learning scheme: $\mathbf{w}(k+1)=\mathbf{w}(k)-\eta(k)\dfrac{\partial Q(k)}{\partial \mathbf{w}}$ | | | |
|---|---|---|---|
| IPLNA error and gradient: | Learning Algorithm (in sketch) | | |
| | NGD (NLMS [13]–[15]) | RLS [19] | ADAM [17] |
| error: $e(k)=y(k)-\tilde{y}(k)$ <br> error criterion: <br> $Q(k)=\dfrac{1}{2}e^2(k)$ <br> gradient: <br> $\dfrac{\partial Q(k)}{\partial \mathbf{w}}=-e(k)\cdot\mathbf{g}(\mathbf{x}(\mathbf{v},k))$ | $\eta(k)=\eta(\mathbf{g}(\mathbf{x}),\mu,k)=$ <br> $=\dfrac{\mu}{\|\mathbf{g}(\mathbf{x})\|_2^2+\varepsilon}$ <br> where the constant learning rate $\mu<2$, and the regularization term $\varepsilon$ is small | $\eta(k)=\eta(\mathbf{g}(\mathbf{x}),\mu,k)=$ <br> $=\mathbf{R}^{-1}(\mathbf{R}^{-1}(k-1),\mathbf{g}(\mathbf{x}),\mu)$ <br> where the constant learning rate is usually $\mu\approx 0.99$. | $\mathbf{w}(k+1)=\mathbf{w}(k)-\eta(k)\mathbf{m}(k)$ <br> $\eta(k)=\dfrac{\mu}{\left(\sqrt{\hat{v}_{(k-1)}}+\varepsilon\right)\cdot(1-\beta_1^k)}$ $\beta_1\approx 0.9$ <br> $\mathbf{m}(k+1)=\beta_1\cdot\mathbf{m}(k)+(1-\beta_1)\dfrac{\partial Q(k)}{\partial \mathbf{w}}$ |

First, let us consider non-momentum gradient methods, such as sketched in Tab. 1 for NLMS or RLS, where the weight-updates for IPLNAs results in time-variant state-space representation as follows

$$\mathbf{w}(k+1)=\left(\mathbf{I}-\eta(k)\cdot\mathbf{g}(\mathbf{x})\cdot\mathbf{g}(\mathbf{x})^T\right)\cdot\mathbf{w}(k)+\eta(k)\cdot y(k)\cdot\mathbf{g}(\mathbf{x}), \quad (2)$$

where $\left(\mathbf{I}-\eta(k)\cdot\mathbf{g}(\mathbf{x})\cdot\mathbf{g}^T(\mathbf{x})\right)$ is the local matrix of dynamics (LMD); $\eta(k)\cdot y(k)\cdot\mathbf{g}(\mathbf{x})$ is the input term, where the time-varying learning rate $\eta(k)$ yields the input gain and $y(k)\cdot\mathbf{g}(\mathbf{x})$ represents the external input term.

Similarly, the time-variant state-space representation can also be obtained for the weight-updates of momentum methods, such as, e.g., for ADAM as sketched in Tab. 1, where the weights $\mathbf{w}$ and their moments $\mathbf{m}$ are mutually connected dynamical systems, and for IPLNA the weight-updates yields

$$\begin{bmatrix}\mathbf{w}(k+1)\\ \mathbf{m}(k+1)\end{bmatrix}=\begin{bmatrix}\mathbf{w}(k)-\eta(k)\mathbf{m}(k)\\ \beta_1\cdot\mathbf{m}(k)+(\beta_1-1)\cdot\mathbf{g}\cdot\mathbf{g}^T\eta(k-1)\mathbf{m}(k-1)\\ -(\beta_1-1)\cdot\mathbf{g}\cdot\mathbf{g}^T\cdot\mathbf{w}(k-1)+(\beta_1-1)\cdot\mathbf{g}\cdot y(k)\end{bmatrix}, \quad (3)$$

where $\mathbf{g}=\mathbf{g}(\mathbf{x},k)$ and (3) can be generally expressed via time-variant state-space representation as follows

$$\boldsymbol{\xi}(k+1)=\mathbf{A}(k)\cdot\boldsymbol{\xi}(k)+\mathbf{B}(k)\cdot\mathbf{u}(k), \quad (4)$$

where the extended state vector $\boldsymbol{\xi}$ consists of weights $\mathbf{w}$ and their momentums $\mathbf{m}$, and the corresponding LMD $\mathbf{A}$, output gain matrix $\mathbf{B}$, and input term $\mathbf{u}$ have changed according to the particular learning rule (see Appendix for details).

For further derivations, it is essential to conclude that the gradient descent learning rule and its existing modifications can be decomposed into some form of in-parameter-linear time-variant state-space representation for IPLNAs, as in (4) for rules with moments (e.g. ADAM) or simpler (e.g. NLMS) as follows

$$\mathbf{w}(k+1)=\mathbf{A}(k)\cdot\mathbf{w}(k)+\mathbf{B}(k)\cdot\mathbf{u}(k), \quad (5)$$

where both $\mathbf{A}(k)$ and $\mathbf{B}(k)$ varies accordingly and are bounded with respect to $k$, and the input vector $\mathbf{u}(k)$ contains measured values. Without loss of generality, $\mathbf{u}(k)$ can also contain measured data at $k+1$, and the state vector of weights can be extended with their momentums, as it results for ADAM in (4)

Thus, without the loss of validity, furher derivations and proofs in this paper adopts learning rule (2) and notation in (5), though the concept is generally valid for IPLNAs including (3), i.e., (4).

## III. BIBS Convergence of IPLNA Gradient Learnings

This section recalls, applies, and proves the BIBS stability concept for a class of incremental weight-update systems and introduces the strict BIBS condition for weight convergence. Based on the proof, a new definition of strict BIBS stability is defined to assure the strict weight convergence of IPLNAs with gradient learning schemes as in (2) or (4), i.e., in a general form (5).

***Definition 2:*** If every bounded input of a system results in a bounded state, then the system is called bounded-input/bounded-state (BIBS) stable. If every bounded input of a system results in a bounded output, then that system is called bounded-input/bounded-output (BIBO) stable.

A system (5) is BIBS stable if there exist two positive constants, such that the conditions

$$\mathbf{w}(k_0)=\mathbf{0}, \|\mathbf{u}(k)\| \leq L_u, \forall k > k_0 \quad (6)$$

imply that $\|\mathbf{w}(k)\| \leq L_w \ \forall k \geq k_0$. The system is BIBO stable if there exists two constants $0 < L_u, L_y < \infty$, such that conditions (6) imply that $|y(k)| \leq L_y \ \forall k \geq k_0$ [12]. □

In practice, the initial weights $\mathbf{w}(k_0)$ are random values; however, it does not violate the BIBS stability as proven further via (7)-(19).

***Theorem 1:*** Time-variant discrete-time weight-update system (5) is BIBS stable if there exist constants $L_u$, $M_A$, and $M_B$ for which $\sup_{k>k_0}\{\|\mathbf{u}(k)\|\}=L_u<\infty$, $\sup_{k>k_0}\{\|\mathbf{A}(k)\|\}=M_A<1$ and $\sup_{k>k_0}\{\|\mathbf{B}(k)\|\}=M_B<\infty$. □

***Proof:*** (based on the proof of Theorem 3 in [12]): The weight-update system (5) unfolds in $k$ via the scheme as follows

$$\mathbf{w}(k)=\mathbf{A}(k-1)\mathbf{w}(k-1)+\mathbf{B}(k-1)\mathbf{u}(k-1), \quad (7)$$

$$\mathbf{w}(k+1)=\mathbf{A}(k)\mathbf{A}(k-1)\mathbf{w}(k-1)+ \\ +\mathbf{A}(k)\mathbf{B}(k-1)\mathbf{u}(k-1)+\mathbf{B}(k)\mathbf{u}(k) \quad (8)$$

$$\vdots$$

$$\mathbf{w}(k+1) = \left[\prod_{j=0}^{k-k_0}\mathbf{A}(k-j)\right]\mathbf{w}(k_0) + \\ + \sum_{i=k_0}^{k-1}\left[\prod_{j=1}^{k-i}\mathbf{A}(k-j+1)\right]\mathbf{B}(i)\mathbf{u}(i) + \mathbf{B}(k)\mathbf{u}(k). \quad (9)$$

Using the notation

$$\mathbb{A}_k=\prod_{j=0}^{k-k_0}\mathbf{A}(k-j), \mathbb{C}_{k,i}=\prod_{j=1}^{k-i}\mathbf{A}(k-j+1), \quad (10)$$

where $i=k_0, k_0+1,\ldots,k-1$, and $\mathbb{C}_{k,k}=\mathbf{I}$ we can rewrite (9) as

$$\mathbf{w}(k+1) = \mathbb{A}_k\mathbf{w}(k_0) + \sum_{i=k_0}^{k}\mathbb{C}_{k,i}\mathbf{B}(i)\mathbf{u}(i). \quad (11)$$

After applying a norm and using the triangle inequality, we obtain

$$\|\mathbf{w}(k+1)\| \leq \|\mathbb{A}_k\|\cdot\|\mathbf{w}(k_0)\|+\sum_{i=k_0}^{k}\|\mathbb{C}_{k,i}\|\cdot\|\mathbf{B}(i)\|\cdot\|\mathbf{u}(i)\| = \mathbb{D}_k. \quad (12)$$

The constraints in Theorem 1 imply the following

$$\|\mathbb{A}_k\| \leq M_A^{k-k_0+1} < 1, \quad (13)$$

$$\|\mathbb{C}_{k,i}\| \leq M_A^{k-i}, \quad (14)$$

$$\|\mathbf{B}(i)\|\cdot\|\mathbf{u}(i)\| \leq M_B \cdot L_u; \quad (15)$$

therefore, it holds for the right-hand side of (12) that

$$\mathbb{D}_k \leq M_A^{k-k_0+1} \cdot \|\mathbf{w}(k_0)\| + M_B \cdot L_u \sum_{i=k_0}^{k} M_A^{k-i}, \quad (16)$$

where the last term represents a partial sum of a geometric sequence, i.e.,

$$\sum_{i=k_0}^{k} M_A^{k-i} = \sum_{i=0}^{k-k_0} M_A^i = \frac{1-M_A^{k-k_0}}{1-M_A} \leq \frac{1}{1-M_A}. \quad (17)$$

Finally, it completes the proof that

$$\|\mathbf{w}(k+1)\| \leq \mathbb{D}_k \leq \|\mathbf{w}(k_0)\| + \frac{M_B L_u}{1-M_B}, \quad (18)$$

where the weight initiation $\mathbf{w}(k_0)$ implies that there exists the constant $L_w = \|\mathbf{w}(k_0)\| + \frac{M_B L_u}{1-M_B}$ such that

$$\|\mathbf{w}(k)\| \leq L_w \text{ for all } k > k_0. \quad \blacksquare \quad (19)$$

## IV. Connotations to Common BIBS and ISS Stability

From the bounded input assumption, it yields that there exist finite upper bounds $M_B, L_u$ for $\|\mathbf{B}(k)\|$ and $\|\mathbf{u}(k)\|$. Using the ratio criterion for convergence of the series in (12), we should find a constant $q<1$ so that

$$\rho(\mathbf{A}(i)) \leq \frac{\|\mathbb{C}_{k,i}\|}{\|\mathbb{C}_{k,i-1}\|} = \|\mathbf{A}(i)\| \leq q, \ i > k_0, \quad (20)$$

where $\rho(\mathbf{A})$ denotes the spectral radius of $\mathbf{A}$.

***Corollary***: IPLNAs with learning-rule state-space representation (2), for which there exists a constant $0 \leq q < 1$ such that

$$\rho\big(\mathbf{I}-\eta(k)\cdot\mathbf{g}(\mathbf{x},k)\cdot\mathbf{g}^T(\mathbf{x},k)\big)\leq q \tag{21}$$

for all $k>k_0$, are BIBS stable. □

**Proof**: It is well known that $\|\mathbf{A}\|<1$ implies $\rho(\mathbf{A})<1$. Recall that for real-time learning $\mathbf{A}(k)=\big(\mathbf{I}-\eta(k)\cdot\mathbf{g}(\mathbf{x},k)\cdot\mathbf{g}^T(\mathbf{x},k)\big)$, $\mathbf{B}(k)=\eta(k)$ and $\mathbf{u}(k)=y(k)\cdot\mathbf{g}(\mathbf{x},k)$ as it is already indicated in (2) or as it can be extended for (4). Therefore, the (21) implies (20); this assures the convergence of the series in (12) and hence

$$\|\mathbf{w}(k)\| \leq \|\mathbf{w}(k_0)\| + K, \tag{22}$$

where $K=M_B L_u /(1-q)$. ∎

**Definition 3**: A system (2) is input-to-state stable (ISS) if there exist two functions $\beta$ and $\gamma$, such that

$$\|\mathbf{w}(k+1)\|\leq\beta\big(\|\mathbf{w}(k_0)\|,k\big)+\gamma\big(\|\mathbf{u}\|\big), \tag{23}$$

where $\beta$ is the $\mathcal{KL}$ function and $\gamma$ is the $\mathcal{K}_\infty$ function [11]. □

**Theorem 2**: IPLNAs that satisfy the condition (20) are ISS. □

**Proof**: From (12), it follows that in the case of (2), we have

$$\|\mathbf{w}(k+1)\|\leq\|\mathbb{A}_k\|\cdot\|\mathbf{w}(k_0)\|+\|\mathbf{u}\|\cdot\sum_{i=k_0}^{k}\|\mathbb{C}_{k,i}\|\cdot\|\mathbf{B}(i)\|, \tag{24}$$

where $\|\mathbf{u}\|=\sup_{k_0\leq i\leq k}\{\|\mathbf{u}(i)\|\}$, then let $\beta\big(\|\mathbf{w}(k_0)\|,k\big)=\|\mathbb{A}_k\|\cdot\|\mathbf{w}(k_0)\|$, and $\gamma\big(\|\mathbf{u}\|\big)=\|\mathbf{u}\|\cdot\sum_{i=k_0}^{k}\|\mathbb{C}_{k,i}\|\cdot\|\mathbf{B}(i)\|$.

From the proof of Corollary 1 it is now clear that $\gamma$ is $\mathcal{K}_\infty$ function.
Now, it is shown that $\beta$ is $\mathcal{KL}$ function.
First, for each $k\geq k_0$ the function $\beta\big(\|\mathbf{w}(k_0)\|,k\big)$ is linear in $\|\mathbf{w}(k_0)\|$, and hence it is of class $\mathcal{K}_\infty$. Second, if the condition (21) holds, then $\lim_{k\to\infty}\mathbb{A}_k=0$. From this follows that $\beta$ belongs into $\mathcal{KL}$ class. ∎

## V. CONSEQUENCES TO REAL APPLICATIONS

In practice, the training data are bounded, and so is the gain matrix $\mathbf{B}(k)$. Thus, the previous sections thoroughly prove that the sufficient condition to maintain the learning convergence under ISS's umbrella and in the sense of BIBS stability is (20) that yields (21) for non-momentum gradient algorithms. For experimental results, e.g., with a class of polynomial neural architectures, please see paper [10], where spectral radii's effect $\rho(\mathbf{A})$ on gradient learning is studied and shown in detail. Also, this letter's proofs are the theoretical complements to the earlier, i.e., more experimentally focused work [10] that did not explicitly show the whole theoretical kinship.

The BIBS condition (20) can be restated as

$$\rho\big(\mathbf{A}(k)\big) \leq \|\mathbf{A}(k)\| \leq 1 \quad \forall k \tag{25}$$

that is very strict, and also, the norm must not necessarily be kept all the time below 1 as it practically fluctuates around 1 as also shown in [10]. Thus, it is more practical to maintain the following condition

$$\frac{\|\mathbb{C}_{k,i}\|}{\|\mathbb{C}_{k,i-p}\|}=\left\|\prod_{j=1}^{k-p}\mathbf{A}(k-j+1)\right\|\leq 1, \tag{26}$$

were $p$ is the custom number of samples, i.e., every $p$-sample sliding window, for which the condition (26) should be maintained. Any unusually large increase of (25) or (26) that extraordinary exceeds 1 than indicates the loss of weight-update stability that has to be avoided, e.g., see Fig. 5 & 6 in [10] as the example of the violation of (20) for a class of polynomial (recurrent) IPLNAs.

## VI. CONCLUSIONS

This paper showed and proved that the ISS framework and BIBS concept are universally applicable to the weight convergence of a broad family of IPLNAs for various modifications of incremental gradient learning algorithms. For real applications, the introduced weight-update stability condition should be monitored and accordingly maintained to avoid the instability of real-time learning systems. The implementation of the weight convergence condition is extraordinarily feasible and very practical; this is because it is enough to monitor the spectral radius that can be practically substituted by calculating the Frobenius norm of the local matrix of dynamics and is achievable in real time on any HW today. Thus, it is also important for autonomous and embedded learning systems implemented by progressive technologies such as FPGA, e.g.

## APPENDIX

The time-variant state-space representation of weight-updates for IPLNAs given in (2) is to be derived, considering (1) with the details in left column of Tab. 1 and respecting the proper vector multiplications, via the steps as follows

$$\tilde{y}(k) = \mathbf{w}(k)^T \cdot \mathbf{g}(\mathbf{x}(k)) = \mathbf{g}(\mathbf{x}(k))^T \cdot \mathbf{w}(k), \tag{27}$$

$$\frac{\partial Q(k)}{\partial \mathbf{w}} = \frac{\partial}{\partial \mathbf{w}}\left(\frac{1}{2}e^2(k)\right) = e(k)\frac{\partial}{\partial \mathbf{w}}\left(y(k) - \mathbf{w}(k)^T \cdot \mathbf{g}(\mathbf{x}(k))\right) = -\mathbf{g}(\mathbf{x}(k)) \cdot e(k), \tag{28}$$

$$\mathbf{w}(k+1) = \mathbf{w}(k) - \eta(k)\frac{\partial Q(k)}{\partial \mathbf{w}} = \mathbf{w}(k) + \eta(k) \cdot \mathbf{g}(\mathbf{x}(k)) \cdot \left(y(k) - \mathbf{g}(\mathbf{x}(k))^T \cdot \mathbf{w}(k)\right) =$$
$$= \mathbf{w}(k) - \eta(k) \cdot \mathbf{g}(\mathbf{x}(k)) \cdot \mathbf{g}(\mathbf{x}(k))^T \cdot \mathbf{w}(k) + \eta(k) \cdot \mathbf{g}(\mathbf{x}(k)) \cdot y(k), \tag{29}$$

that returns in the form of (2) as follows

$$\mathbf{w}(k+1) = \left(\mathbf{I} - \eta(k) \cdot \mathbf{g}(\mathbf{x}(k)) \cdot \mathbf{g}(\mathbf{x}(k))^T\right) \cdot \mathbf{w}(k) + \eta(k) \cdot y(k) \cdot \mathbf{g}(\mathbf{x}(k)). \tag{2}$$

State-space representation of the weight-update system for ADAM leading to (3) and (4) derives as follows

$$\mathbf{w}(k+1) = \mathbf{w}(k) - \eta(k) \cdot \mathbf{m}(k), \tag{30}$$

$$\mathbf{m}(k+1) = \beta_1 \cdot \mathbf{m}(k) + (1 - \beta_1) \cdot \frac{\partial Q(k)}{\partial \mathbf{w}} =$$
$$= \beta_1 \cdot \mathbf{m}(k) + (\beta_1 - 1) \cdot \mathbf{g}(\mathbf{x},k) \cdot e(k) =$$
$$= \beta_1 \cdot \mathbf{m}(k) + (\beta_1 - 1) \cdot \mathbf{g}(\mathbf{x},k) \cdot \left(y(k) - \mathbf{g}(\mathbf{x},k)^T \cdot \mathbf{w}(k)\right) =$$
$$= \beta_1 \cdot \mathbf{m}(k) - (\beta_1 - 1) \cdot \mathbf{g}(\mathbf{x},k) \cdot \mathbf{g}(\mathbf{x},k)^T \cdot \mathbf{w}(k) + (\beta_1 - 1) \cdot \mathbf{g}(\mathbf{x},k) \cdot y(k), \tag{31}$$

where substituting step-delayed (30) into (31) leads to

$$\mathbf{m}(k+1) = \beta_1 \cdot \mathbf{m}(k) - (\beta_1 - 1)\mathbf{g}(\mathbf{x},k)\mathbf{g}(\mathbf{x},k)^T \left(\mathbf{w}(k-1) - \eta(k-1) \cdot \mathbf{m}(k-1)\right) + (\beta_1 - 1)\mathbf{g}(\mathbf{x},k)y(k), \tag{32}$$

$$\mathbf{m}(k+1) = \beta_1 \cdot \mathbf{m}(k) + (\beta_1 - 1)\mathbf{g}(\mathbf{x},k)\mathbf{g}(\mathbf{x},k)^T \eta(k-1)\mathbf{m}(k-1) - (\beta_1 - 1)\mathbf{g}(\mathbf{x},k)\mathbf{g}(\mathbf{x},k)^T \mathbf{w}(k-1) + (\beta_1 - 1)\mathbf{g}(\mathbf{x},k)y(k), \tag{33}$$

so the state-space representation of IPLNA learning rule with moments can be here rewritten in the following matrix form

$$\begin{bmatrix} \mathbf{w}(k) \\ \mathbf{w}(k+1) \\ \mathbf{m}(k) \\ \mathbf{m}(k+1) \end{bmatrix} = \mathbf{A}(k) \begin{bmatrix} \mathbf{w}(k-1) \\ \mathbf{w}(k) \\ \mathbf{m}(k-1) \\ \mathbf{m}(k) \end{bmatrix} + \begin{bmatrix} 0 \\ 0 \\ 0 \\ (\beta_1 - 1) \end{bmatrix} \cdot \big[ \mathbf{g}(\mathbf{x},k) \cdot y(k) \big], \tag{34}$$

where $\mathbf{A}(k)$ is in detail for ADAM as follows

$$\mathbf{A}(k) = \begin{bmatrix} 0 & 1 & 0 & 0 \\ 0 & 1 & 0 & -\eta(k) \\ 0 & 0 & 0 & 1 \\ -(\beta_1 - 1)\mathbf{g}(\mathbf{x},k)\mathbf{g}(\mathbf{x},k)^T & 0 & (\beta_1 - 1)\mathbf{g}(\mathbf{x},k)\mathbf{g}(\mathbf{x},k)^T \eta(k-1) & \beta_1 \end{bmatrix}, \tag{35}$$

so the ADAM weight-update system (30) & (31) turns into the already shown general state-space representation for IPLNAs as

$$\boldsymbol{\xi}(k+1) = \mathbf{A}(k) \cdot \boldsymbol{\xi}(k) + \mathbf{B}(k) \cdot \mathbf{u}(k), \tag{4}$$

where the state vector $\boldsymbol{\xi}$, local matrix of dynamics $\mathbf{A}$, output gain matrix (or vector) $\mathbf{B}$, and input term $\mathbf{u}$ changes accordingly. Therefore, the further validity and application of BIBS stability applied to the weight convergence are valid and straightforward.